\documentclass[runningheads]{llncs}
\usepackage[T1]{fontenc}
\usepackage{graphicx,verbatim}
\usepackage{booktabs}
\usepackage{multirow}
\usepackage{amsmath}
\usepackage{amssymb}
\usepackage{adjustbox}
\usepackage{makecell}
\usepackage{float}
\usepackage{subcaption}
\usepackage[table,xcdraw]{xcolor}
\usepackage{xspace}
\usepackage[hidelinks]{hyperref}
\newcommand{\secondbest}[1]{\underline{#1}}

\begin{document}

\title{Hierarchical Perfusion Graphs for Tumor Heterogeneity Modeling in Glioma Molecular Subtyping}

\titlerunning{HiPerfGNN}

\author{Han Jang$^*$\inst{1} \and
Junhyeok Lee$^*$\inst{2} \and
Heeseong Eum\inst{2} \and
Joon Jang\inst{3} \and
Yoseob Han\inst{4} \and
Seung Hong Choi\inst{1,2,5,6} \and
Kyu Sung Choi$^{\dagger}$\inst{5,6,7}}

\authorrunning{H. Jang and J. Lee et al.}
\institute{
Interdisciplinary Program in Bioengineering, Seoul National University, Seoul, South Korea \and
Interdisciplinary Program in Cancer Biology, Seoul National University College of Medicine, Seoul, South Korea \and
Dept. of Biomedical Sciences, Seoul National University, Seoul, South Korea \and
Dept. of Electronic Engineering, Soongsil University, Seoul, South Korea \and
Dept. of Radiology, Seoul National University Hospital, Seoul, South Korea \and
Dept. of Radiology, Seoul National University College of Medicine, Seoul, South Korea \and
Healthcare AI Research Institute, Seoul National University Hospital, Seoul, South Korea\\
\email{ent1127@snu.ac.kr}
}

\maketitle
\renewcommand{\thefootnote}{}
\footnotetext{$^*$ Equal contribution. $^\dagger$ Corresponding authors.}
\renewcommand{\thefootnote}{\arabic{footnote}}

\begin{abstract}

Precise molecular subtyping of gliomas, including isocitrate dehydrogenase~(IDH) mutation and 1p/19q codeletion, directly guides surgical and therapeutic decisions, yet currently relies on invasive tissue sampling.
Deep learning on structural MRI has emerged as a non-invasive alternative, but anatomy-only approaches cannot capture the hemodynamic signatures that distinguish molecular subtypes.
Radiogenomics based on dynamic susceptibility contrast~(DSC) MRI holds immense potential for non-invasively characterizing glioma molecular subtypes, yet clinical deployment has been hindered by inter-site variability and the limitations of voxel-wise analysis.
We introduce HiPerfGNN, a framework that first learns discrete hemodynamic representations from raw time-intensity curves using a vector-quantized variational autoencoder~(VQ-VAE).
These quantized perfusion codes define coarse-level graph nodes representing functional tumor habitats, each of which is hierarchically subdivided into fine-level subregions guided by structural MRI.
A hierarchical graph neural network then propagates information across scales for molecular prediction.
On an internal cohort~($n{=}475$), the model achieved AUCs of 0.96 (IDH), 0.89 (1p/19q), and 0.84 (WHO grade), and maintained robust IDH performance~(AUC 0.89) on an independent external cohort~($n{=}397$; IDH labels only) without recalibration.
Gradient-based saliency analysis confirms biologically grounded attention patterns aligned with known glioma pathophysiology.
Our results demonstrate the added value of integrating perfusion dynamics into radiogenomic pipelines for glioma molecular subtyping.
Code is available at \url{https://github.com/janghana/HiPerfGNN}.

\keywords{Glioma molecular subtyping \and Perfusion MRI \and Hierarchical graph neural network \and Vector quantization \and Radiogenomics}

\end{abstract}

\section{Introduction}

Accurate molecular characterization of gliomas is essential for guiding surgical planning, adjuvant therapy, and prognostication~\cite{wen2008malignant}.
The 2021 World Health Organization~(WHO) classification of Central Nervous System~(CNS) tumors formalizes this by defining glioma subtypes primarily through molecular markers, including isocitrate dehydrogenase~(IDH) mutation and 1p/19q codeletion~\cite{louis20212021}.
However, confirmation of these markers currently requires invasive tissue sampling through biopsy or resection, procedures that carry substantial risks of neurological morbidity for tumors in eloquent brain regions~\cite{weller2017european}, highlighting the urgent need for reliable non-invasive alternatives~\cite{smits2021mri}.
Radiogenomics has emerged to address this gap, aiming to decode molecular characteristics from diverse modalities including structural, functional, and perfusion MRI~\cite{aerts2014decoding,gillies2016radiomics}.

Deep learning applied to structural MRI has shown promise in predicting molecular genotypes~\cite{chang2018residual,choi2021fully,farahani2025foundbionet}.
Among structural methods, GlioMT~\cite{byeon2025interpretable} demonstrated that multimodal transformers fusing multiple structural MRI sequences can achieve strong IDH prediction performance by capturing complementary anatomical contrasts.
Nevertheless, even such multi-sequence structural approaches provide limited
insight into the physiological mechanisms linking genotype and phenotype.
Critically, IDH mutations directly alter tumor angiogenesis, with IDH-mutant gliomas exhibiting a relatively normalized vasculature, whereas IDH-wildtype tumors develop disorganized, hypervascular networks driven by upregulated hypoxia-inducible factor~(HIF-1$\alpha$) signaling~\cite{leu2017perfusion,kickingereder2015idh}.
This hemodynamic divergence motivates the use of dynamic susceptibility contrast~(DSC) perfusion MRI, which captures contrast agent dynamics through the tumor vasculature to probe neovascularization~\cite{law2008gliomas,choi2019prediction}.
These spatiotemporal perfusion patterns provide sensitive readouts of microvascular proliferation and vascular leakage that structural imaging cannot directly measure, yet fully exploiting this rich information remains analytically challenging~\cite{artzi2019differentiation}.

Early analytical frameworks simplified DSC time-intensity curves~(TICs) into static parametric maps such as relative cerebral blood volume~(rCBV), yet such derived maps are inherently limited in modelling complex physiological kinetics and inevitably discard important temporal dynamics~\cite{anzalone2018brain,kickingereder2016radiomic}.
Subsequent advances incorporated recurrent neural networks~(RNNs) to directly process raw temporal signals~\cite{choi2019prediction}.
More recently, PerfGAT introduced graph attention networks for perfusion-based IDH prediction~\cite{yan2024spatiotemporal}, but its reliance on predefined brain atlases originally developed for healthy brains imposes rigid boundaries that fail to adapt to tumor-induced deformations~\cite{zacharaki2009classification,bakas2018identifying}.
These limitations motivate a shift from voxel-centric modeling toward entity-based representations.
Graph Neural Networks~(GNNs) naturally support this paradigm by encoding tumor subregions as nodes and their relationships as edges, enabling explicit modeling of intratumoral heterogeneity~\cite{pati2022hierarchical,kipf2016semi,yan2024spatiotemporal}.

In this work, we present HiPerfGNN, to the best of our knowledge, the first radiogenomic framework that jointly integrates structural and perfusion MRI into tumor-specific hierarchical graph representations learned directly from perfusion dynamics using a VQ-VAE~\cite{van2017neural}.
Our contributions are as follows:
(1)~a tumor-specific hierarchical graph framework that learns topology from DSC perfusion dynamics, capturing both coarse-grained functional clusters and fine-grained anatomical subregions;
(2)~an unsupervised VQ-VAE strategy to encode raw TICs into robust hemodynamic codes that generalize across acquisition protocols;
(3)~comprehensive evaluation for IDH, 1p/19q, and WHO grade prediction with external validation demonstrating cross-site generalization.

\begin{figure}[t]
\centering
\includegraphics[width=\textwidth]{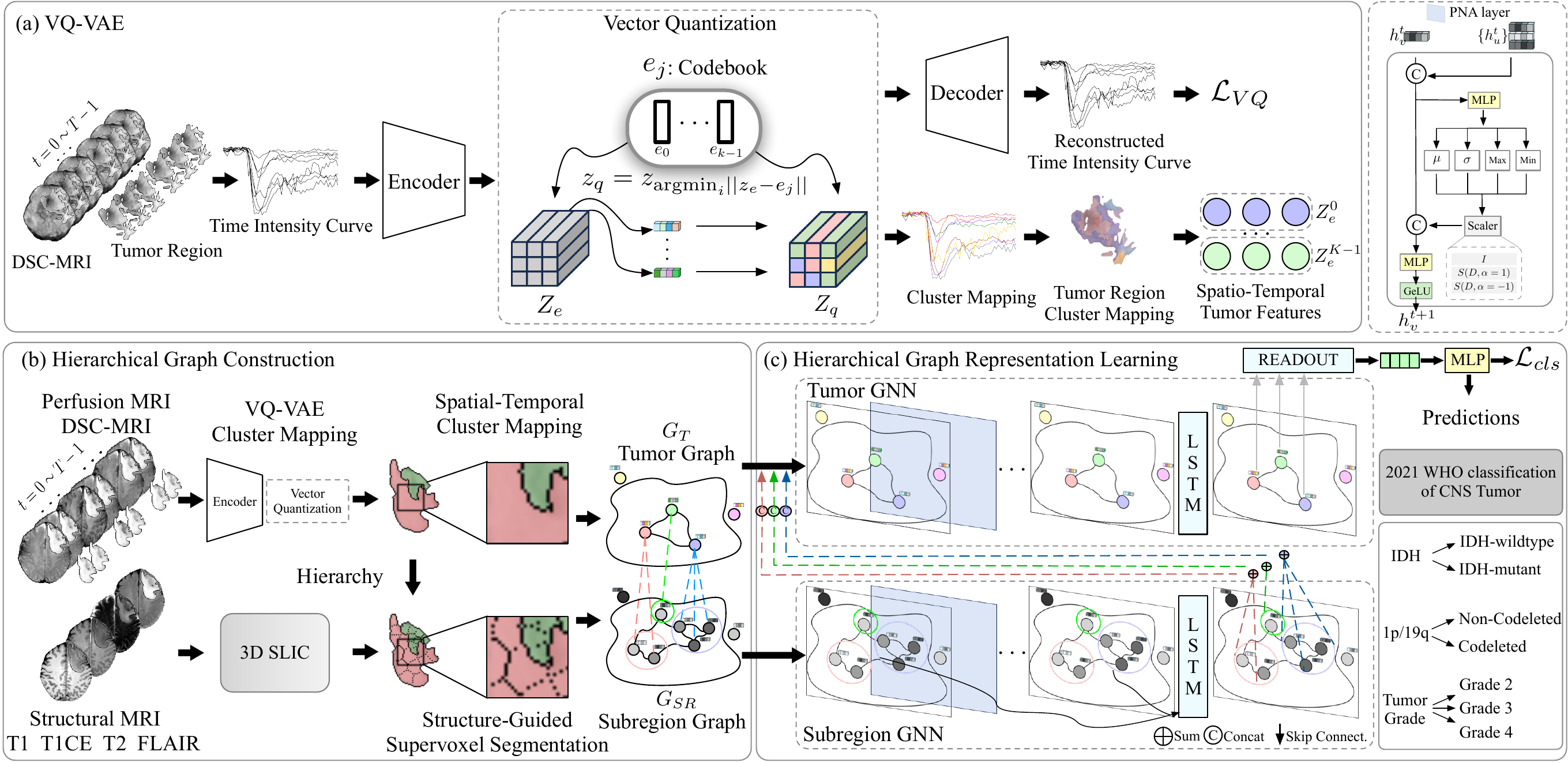}
\caption{
\textbf{Architecture overview.}
(a)~VQ-VAE: encoding, quantization, and cluster mapping of DSC-MRI time-intensity curves.
(b)~Hierarchical graph construction from perfusion features and structure-guided supervoxels.
(c)~Fine-to-coarse GNN with multi-scale readout for molecular classification.
}
\label{fig:architecture}
\end{figure}

\section{Method}

As illustrated in Fig.~\ref{fig:architecture}, our framework consists of three stages:
unsupervised perfusion quantization via VQ-VAE,
hierarchical graph construction integrating perfusion and structural information,
and hierarchical graph learning for molecular prediction.

\subsection{Unsupervised Perfusion Quantization via VQ-VAE}

Let $I{\in}\mathbb{R}^{H\!\times\!W\!\times\!D\!\times\!T}$ denote the 4D DSC-MRI data and $M{\in}\{0,1\}^{H\!\times\!W\!\times\!D}$ the tumor mask (Fig.~\ref{fig:architecture}a).
For each voxel within $M$, a 1D time-intensity curve $x{\in}\mathbb{R}^T$ is extracted, temporally z-scored, and fed into a VQ-VAE~\cite{van2017neural} that handles variable acquisition lengths~($T{=}45$--$60$) via adaptive pooling.

The encoder maps $x$ to a fixed-size latent sequence $Z_e(x){\in}\mathbb{R}^{N\times d_\text{enc}}$.
Each latent vector $z_e^{(n)}$ is quantized to its nearest codebook entry $e_{k_n}$ from a learnable codebook $\mathcal{E}{=}\{e_k\}_{k=1}^K$:
\begin{equation}
k_n = \arg\min_{j}\|z_e^{(n)}(x) - e_j\|_2^2,\quad n=1,\dots,N.
\label{eq:quantization}
\end{equation}
The composite training objective combines $\ell_1$ reconstruction with commitment and codebook losses:
\begin{equation}
\mathcal{L}_\text{VQ} = \|x - \hat{x}\|_1 + \sum_{n=1}^{N}\!\Big(\|\text{sg}[z_e^{(n)}] - z_q^{(n)}\|_2^2 + \beta\|z_e^{(n)} - \text{sg}[z_q^{(n)}]\|_2^2\Big),
\label{eq:vq_loss}
\end{equation}
where $\text{sg}[\cdot]$ is the stop-gradient operator and $\beta$ controls commitment strength.
With codebook size $K{=}2$ and $N{=}3$ latent vectors per TIC (yielding $K^N{=}8$ possible discrete code sequences per voxel), the model discovers distinct hemodynamic patterns, capturing the tri-phasic kinetics of DSC-MRI (baseline, first-pass drop, recirculation) while enforcing regularization for cross-site robustness.

\subsection{Hierarchical Tumor Graph Construction}
\label{sec:graph_construction}

The hierarchical graph $G_H{=}\{G_T, G_{SR}, A_{SR\to T}\}$ encodes tumor information at two complementary scales (Fig.~\ref{fig:architecture}b).

\paragraph{Coarse-level Tumor Graph ($G_T$).}
A 3D label map $L$ is generated by assigning each tumor voxel its discrete code sequence from Eq.~\eqref{eq:quantization}.
Spatially contiguous voxels sharing identical codes form connected components, each becoming a node $v{\in}V_T$ representing a hemodynamic habitat.
The node feature $h_v{\in}\mathbb{R}^{d_\text{flat}}$ ($d_\text{flat}{=}N{\times}d_\text{enc}$) is computed as the mean of flattened encoder latents over constituent voxels:
\begin{equation}
h_v = \frac{1}{|\mathcal{V}_v|}\sum_{x\in\mathcal{V}_v}\text{vec}(Z_e(x)).
\label{eq:coarse_node}
\end{equation}
Edges $E_T$ connect nodes via $k$-nearest-neighbor ($k$NN) graph construction on centroid distances, pruned by a spatial threshold $\delta_\text{max}$.
Edge features $w_{uv}{\in}\mathbb{R}^{N^2}$ are flattened cosine-similarity matrices between the latent sequences of connected nodes.

\paragraph{Fine-level Subregion Graph ($G_{SR}$).}
Within each coarse node's spatial domain, we apply 3D SLIC supervoxelization~\cite{achanta2012slic} on the co-registered four-channel structural MRI volume $S{\in}\mathbb{R}^{H\!\times\!W\!\times\!D\!\times\!C}$ ($C{=}4$; T1, T1CE, T2, FLAIR).
Each supervoxel becomes a fine-level node $v_{SR}{\in}V_{SR}$ with features $h_{v_{SR}}{\in}\mathbb{R}^C$ computed as the mean voxel intensity per modality.
Edges follow the same $k$NN construction, with $C^2$-dimensional features from the cross-channel similarity matrix.
This structure-guided subdivision resolves structurally homogeneous subregions within each perfusion-defined habitat.

\paragraph{Hierarchical Assignment.}
A sparse binary matrix $A_{SR\to T}{\in}\{0,1\}^{|V_{SR}|\times|V_T|}$ links each fine-level node to its unique parent coarse-level node based on spatial containment, ensuring a strict many-to-one mapping.

\subsection{Hierarchical Graph Neural Network}
\label{sec:hgnn}
The HGNN follows a fine-to-coarse learning strategy (Fig.~\ref{fig:architecture}c).
Both levels employ Principal Neighborhood Aggregation~(PNA)~\cite{corso2020principal} layers with multiple aggregators (mean, std, max, min) and degree-scalers, enabling the model to distinguish diverse neighborhood topologies.

\paragraph{Fine-level Learning.}
A dedicated GNN $\mathcal{G}_{SR}$ processes $G_{SR}$ through $T_{SR}$ PNA layers.
To capture multi-scale receptive fields and prevent over-smoothing, an LSTM-based Jumping Knowledge~(JK) readout~\cite{xu2018representation} aggregates intermediate representations:
\begin{equation}
h_{v_{SR}}^\text{final} = \text{LSTM}\!\left(\{h_{v_{SR}}^{(t)}\}_{t=1}^{T_{SR}}\right).
\label{eq:jk_sr}
\end{equation}

\paragraph{Coarse-level Enrichment.}
Fine-level knowledge is propagated upward via $A_{SR\to T}$.
Each coarse node's feature is enriched by concatenating its perfusion descriptor with mean-pooled anatomical embeddings from child supervoxels:
\begin{equation}
h_{v_T}^{(0)} = \text{Concat}\!\left(h_{v_T}^\text{init},\;\frac{1}{|\mathcal{C}(v_T)|}\sum_{v_{SR}\in\mathcal{C}(v_T)}h_{v_{SR}}^\text{final}\right).
\label{eq:coarse_enrich}
\end{equation}
The coarse GNN $\mathcal{G}_T$ then processes the enriched tumor graph through $T_T$ PNA layers with an analogous JK-LSTM readout, followed by global mean pooling and an MLP classifier.

\paragraph{Training Objective.}
Training minimizes a weighted cross-entropy loss for prediction target~(IDH, 1p/19q, WHO grade):
$\mathcal{L}_\text{cls} = \sum_{c}w_c\cdot\text{CE}(y_c,\hat{y}_c)$.

\paragraph{Gradient-based Saliency.}
For clinical interpretability, we compute gradient-based node importance scores~\cite{pope2019explainability} for each coarse node $v{\in}V_T$.
Each node score is then distributed to its constituent voxels defined by the label map $L$ and spatially smoothed via Gaussian filtering, producing voxel-level saliency maps overlaid on the co-registered structural and DSC-MRI volumes.

\section{Experiments and Results}

\subsection{Datasets and Setup}

\paragraph{Datasets.}
For the internal cohort ($n{=}475$), patients with diffuse gliomas (mean age $55.3{\pm}14.5$ years; 52.8\% male) were retrospectively collected from a single center (IRB-approved, consent waived).
Each patient had structural MRI (T1, T1CE, T2, FLAIR) and DSC perfusion MRI, with tumor masks generated by HD-GLIO~\cite{kickingereder2019automated} and validated by an expert neuroradiologist, and molecular profiling per 2021 WHO criteria.
The cohort comprised 339 IDH-wildtype (71.4\%) and 136 IDH-mutant (28.6\%) cases, of which 70 (14.7\%) harbored 1p/19q codeletion.
WHO grade distribution was Grade~4 ($n{=}350$, 73.7\%), Grade~3 ($n{=}78$, 16.4\%), and Grade~2 ($n{=}47$, 9.9\%).
Data were split at the patient level into training ($n{=}379$), validation ($n{=}48$), and test ($n{=}48$) sets with stratified molecular distributions.
For external validation, the publicly available University of Pennsylvania Glioblastoma dataset~(UPenn-GBM, $n{=}397$)~\cite{bakas2022university} (mean age $62.4{\pm}11.9$ years; 58.7\% male) consisted exclusively of WHO Grade~4 glioblastomas with 386 IDH-wildtype (97.2\%) and only 11 IDH-mutant (2.8\%) cases; 1p/19q status was unavailable.
No retraining or recalibration was performed.

\paragraph{Implementation.}
The framework was implemented in PyTorch~2.5.1 and PyG~2.6.1.
The VQ-VAE ($K{=}2$, $N{=}3$, $d_\text{enc}{=}256$) was trained exclusively on training and validation patients for 100 epochs (Adam, lr=$10^{-5}$, batch=512) and frozen; all test and external patients were excluded from codebook learning.The HGNN ($T_{SR}{=}T_T{=}3$ PNA layers, $k{=}5$, $\delta_\text{max}{=}15$~voxels, two-layer MLP classifier) was trained for 30 epochs (AdamW, lr=$10^{-3}$, weight decay=$10^{-4}$) with early stopping (patience=10), GraphNorm and dropout~(0.3)~\cite{cai2021graphnorm}.
All experiments used two NVIDIA RTX 3090 GPUs (24\,GB each).

\begin{table}[t]
\centering
\caption{
\textbf{Comparative performance of radiogenomic models.}
Performance metrics with 95\% confidence intervals from 1000-iteration stratified bootstrapping. Multi-class AUC is macro-averaged one-vs-rest.
\textbf{Bold} indicates best; \secondbest{underline} indicates second best.
}
\label{tab:main}
\resizebox{\textwidth}{!}{%
\begin{tabular}{llcccccc}
\toprule
Dataset & Task & Model & Accuracy & F1 (macro) & Sensitivity & Specificity & AUC \\
\midrule
\multirow{9}{*}{\makecell[cl]{Internal\\test}}
 & \multirow{3}{*}{IDH}
 & PerfGAT & 0.71\,(0.58--0.83) & \secondbest{0.72}\,(0.58--0.84) & \secondbest{0.71}\,(0.58--0.83) & 0.72\,(0.56--0.87) & 0.699\,(0.515--0.865) \\
 & & GlioMT  & \secondbest{0.73}\,(0.60--0.85) & 0.42\,(0.15--0.68) & 0.31\,(0.10--0.56) & \secondbest{0.84}\,(0.74--0.94) & \secondbest{0.858}\,(0.739--0.954) \\
 & & HiPerfGNN    & \textbf{0.88}\,(0.76--0.97) & \textbf{0.84}\,(0.70--0.98) & \textbf{0.98}\,(0.97--1.00) & \textbf{0.88}\,(0.76--0.97) & \textbf{0.963}\,(0.887--1.000) \\
\cmidrule(lr){2-8}
 & \multirow{3}{*}{1p/19q}
 & PerfGAT & 0.56\,(0.42--0.71) & 0.61\,(0.46--0.73) & 0.56\,(0.42--0.71) & 0.51\,(0.35--0.67) & 0.647\,(0.469--0.833) \\
 & & GlioMT  & \secondbest{0.75}\,(0.50--0.94) & \secondbest{0.75}\,(0.47--0.94) & \secondbest{0.76}\,(0.50--0.94) & \secondbest{0.58}\,(0.21--0.80) & \secondbest{0.714}\,(0.383--0.980) \\
 & & HiPerfGNN    & \textbf{0.88}\,(0.79--0.97) & \textbf{0.86}\,(0.73--0.97) & \textbf{0.86}\,(0.75--0.94) & \textbf{0.98}\,(0.87--1.00) & \textbf{0.887}\,(0.789--0.979) \\
\cmidrule(lr){2-8}
 & \multirow{3}{*}{\makecell[cl]{WHO\\Grade}}
 & PerfGAT & 0.15\,(0.06--0.25) & 0.09\,(0.04--0.15) & 0.33\,(0.33--0.53) & 0.67\,(0.66--0.67) & 0.657\,(0.531--0.786) \\
 & & GlioMT  & \secondbest{0.69}\,(0.56--0.81) & \secondbest{0.38}\,(0.27--0.51) & \secondbest{0.39}\,(0.28--0.53) & \secondbest{0.75}\,(0.67--0.84) & \secondbest{0.773}\,(0.603--0.915) \\
 & & HiPerfGNN    & \textbf{0.83}\,(0.71--0.92) & \textbf{0.65}\,(0.45--0.86) & \textbf{0.66}\,(0.49--0.86) & \textbf{0.90}\,(0.83--0.97) & \textbf{0.840}\,(0.678--0.965) \\
\midrule
\multirow{3}{*}{\makecell[cl]{External\\(UPenn)}}
 & \multirow{3}{*}{IDH}
 & PerfGAT & 0.59\,(0.54--0.64) & 0.08\,(0.04--0.14) & \secondbest{0.64}\,(0.33--0.91) & 0.58\,(0.54--0.63) & 0.565\,(0.387--0.717) \\
 & & GlioMT  & \secondbest{0.88}\,(0.86--0.92) & \secondbest{0.38}\,(0.20--0.67) & 0.27\,(0.12--0.55) & \textbf{0.89}\,(0.82--0.98) & \secondbest{0.857}\,(0.735--0.968) \\
 & & HiPerfGNN    & \textbf{0.91}\,(0.80--0.98) & \textbf{0.93}\,(0.84--0.99) & \textbf{0.74}\,(0.57--0.89) & \secondbest{0.87}\,(0.73--0.96) & \textbf{0.887}\,(0.797--0.989) \\
\bottomrule
\end{tabular}}
\end{table}

\subsection{Competing Methods}
We compared against two state-of-the-art approaches, namely
PerfGAT~\cite{yan2024spatiotemporal}, a perfusion-only graph attention network using atlas-based parcellation (SRI24/TZO), and
GlioMT~\cite{byeon2025interpretable}, a structure-only multimodal transformer fusing four structural MRI sequences.
Both were implemented per published protocols.
This selection enables evaluation across perfusion-only vs.\ structure-only paradigms, fixed atlas-based vs.\ tumor-specific graphs, and single- vs.\ multi-modal strategies.

\subsection{Classification Performance}
Table~\ref{tab:main} summarizes the main results.
On the internal test set, HiPerfGNN achieved superior performance across all three tasks, with IDH AUC of 0.963 (balanced accuracy: 0.93) vs.\ 0.858 for GlioMT and 0.699 for PerfGAT, 1p/19q AUC of 0.887 vs.\ 0.714 and 0.647, and WHO grade AUC of 0.840 vs.\ 0.773 and 0.657.

The advantage was most pronounced on external validation.
Without any recalibration, HiPerfGNN maintained an IDH AUC of 0.887 on UPenn-GBM, whereas GlioMT reached 0.857 and PerfGAT degraded to 0.565.
Importantly, HiPerfGNN also achieved the highest F1-score of 0.93 and accuracy of 0.91 on the external set, confirming that the learned representations capture genotype-specific hemodynamic signatures rather than simply exploiting grade and genotype correlations, as the external cohort consists exclusively of Grade~4 tumors.

\begin{table}[!t]
\centering
\caption{
\textbf{Ablation study comparing graph architecture variants.}
95\% CI from 1000-iteration stratified bootstrapping. Multi-class AUC is macro-averaged one-vs-rest.
\textbf{Bold} indicates best; \secondbest{underline} indicates second best.
}
\label{tab:ablation}
\resizebox{\textwidth}{!}{%
\begin{tabular}{ll ccc ccccc}
\toprule
Dataset & Task & Perf & Hier & SG & Accuracy & F1 (macro) & Sensitivity & Specificity & AUC \\
\midrule
\multirow{12}{*}{\makecell[cl]{Internal\\test}}
 & \multirow{4}{*}{IDH}
 & \checkmark & & & 0.73\,(0.60--0.85) & 0.55\,(0.30--0.75) & 0.51\,(0.25--0.75) & 0.84\,(0.69--0.96) & 0.748\,(0.570--0.901) \\
 & & \checkmark & \checkmark & & 0.77\,(0.65--0.88) & 0.58\,(0.35--0.78) & 0.50\,(0.27--0.76) & \textbf{0.91}\,(0.80--1.00) & 0.785\,(0.631--0.927) \\
 & & & & \checkmark & \secondbest{0.87}\,(0.77--0.96) & \secondbest{0.80}\,(0.62--0.94) & \secondbest{0.81}\,(0.60--1.00) & \textbf{0.91}\,(0.79--1.00) & \secondbest{0.875}\,(0.753--0.964) \\
 & & \checkmark & \checkmark & \checkmark & \textbf{0.88}\,(0.76--0.97) & \textbf{0.84}\,(0.70--0.98) & \textbf{0.98}\,(0.97--1.00) & \secondbest{0.88}\,(0.76--0.97) & \textbf{0.963}\,(0.887--1.000) \\
\cmidrule(lr){2-10}
 & \multirow{4}{*}{1p/19q}
 & \checkmark & & & 0.81\,(0.71--0.92) & 0.39\,(0.24--0.53) & 0.38\,(0.27--0.53) & 0.95\,(0.86--0.98) & 0.641\,(0.418--0.867) \\
 & & \checkmark & \checkmark & & \secondbest{0.85}\,(0.75--0.94) & \secondbest{0.83}\,(0.71--0.95) & \secondbest{0.73}\,(0.50--0.88) & \secondbest{0.97}\,(0.91--1.00) & 0.766\,(0.601--0.902) \\
 & & & & \checkmark & 0.83\,(0.73--0.92) & 0.82\,(0.69--0.94) & 0.34\,(0.01--0.67) & 0.95\,(0.87--0.99) & \secondbest{0.840}\,(0.714--0.940) \\
 & & \checkmark & \checkmark & \checkmark & \textbf{0.88}\,(0.79--0.97) & \textbf{0.86}\,(0.73--0.97) & \textbf{0.74}\,(0.68--0.88) & \textbf{0.98}\,(0.87--1.00) & \textbf{0.887}\,(0.789--0.979) \\
\cmidrule(lr){2-10}
 & \multirow{4}{*}{\makecell[cl]{WHO\\Grade}}
 & \checkmark & & & 0.65\,(0.50--0.77) & 0.43\,(0.30--0.60) & 0.46\,(0.30--0.65) & 0.67\,(0.55--0.80) & 0.673\,(0.416--0.738) \\
 & & \checkmark & \checkmark & & 0.67\,(0.54--0.80) & 0.66\,(0.52--0.79) & \secondbest{0.69}\,(0.54--0.80) & 0.63\,(0.42--0.83) & 0.721\,(0.672--0.779) \\
 & & & & \checkmark & \secondbest{0.77}\,(0.66--0.89) & \secondbest{0.79}\,(0.68--0.90) & 0.77\,(0.66--0.89) & \secondbest{0.86}\,(0.70--0.98) & \secondbest{0.839}\,(0.732--0.950) \\
 & & \checkmark & \checkmark & \checkmark & \textbf{0.84}\,(0.74--0.94) & \textbf{0.83}\,(0.73--0.94) & \textbf{0.84}\,(0.74--0.94) & \textbf{0.90}\,(0.82--0.97) & \textbf{0.839}\,(0.681--0.972) \\
\midrule
\multirow{4}{*}{\makecell[cl]{External\\(UPenn)}}
 & \multirow{4}{*}{IDH}
 & \checkmark & & & 0.75\,(0.71--0.80) & 0.84\,(0.80--0.87) & 0.27\,(0.00--0.55) & 0.77\,(0.72--0.81) & 0.633\,(0.431--0.782) \\
 & & \checkmark & \checkmark & & 0.69\,(0.64--0.74) & 0.56\,(0.51--0.61) & \secondbest{0.73}\,(0.68--0.77) & 0.55\,(0.50--0.61) & \secondbest{0.699}\,(0.583--0.781) \\
 & & & & \checkmark & \secondbest{0.87}\,(0.84--0.91) & \secondbest{0.90}\,(0.88--0.94) & 0.55\,(0.49--0.60) & \textbf{0.88}\,(0.85--0.92) & 0.689\,(0.660--0.856) \\
 & & \checkmark & \checkmark & \checkmark & \textbf{0.91}\,(0.88--0.94) & \textbf{0.93}\,(0.90--0.95) & \textbf{0.74}\,(0.69--0.78) & \secondbest{0.87}\,(0.84--0.91) & \textbf{0.887}\,(0.810--0.956) \\
\bottomrule
\end{tabular}}
\end{table}

\subsection{Ablation Study}
We systematically evaluated three architectural components of HiPerfGNN (Table~\ref{tab:ablation}): perfusion quantization (Perf), hierarchical graph construction (Hier), and structure-guided subdivision (SG).

On the internal set, adding hierarchical modeling to perfusion alone improved IDH AUC from 0.748 to 0.785. The structure-only variant achieved a competitive IDH AUC of 0.875. The full model further boosted performance, achieving IDH AUC of 0.963, 1p/19q AUC of 0.887, and Grade AUC of 0.839.
On external validation, both single-component models degraded substantially (perfusion-only: 0.633, structure-only: 0.689), while HiPerfGNN maintained an AUC of 0.887. This confirms that neither modality alone generalizes across sites, and their hierarchical integration is key to robust cross-cohort performance.

\begin{figure}[h!]
\centering
\includegraphics[width=\textwidth]{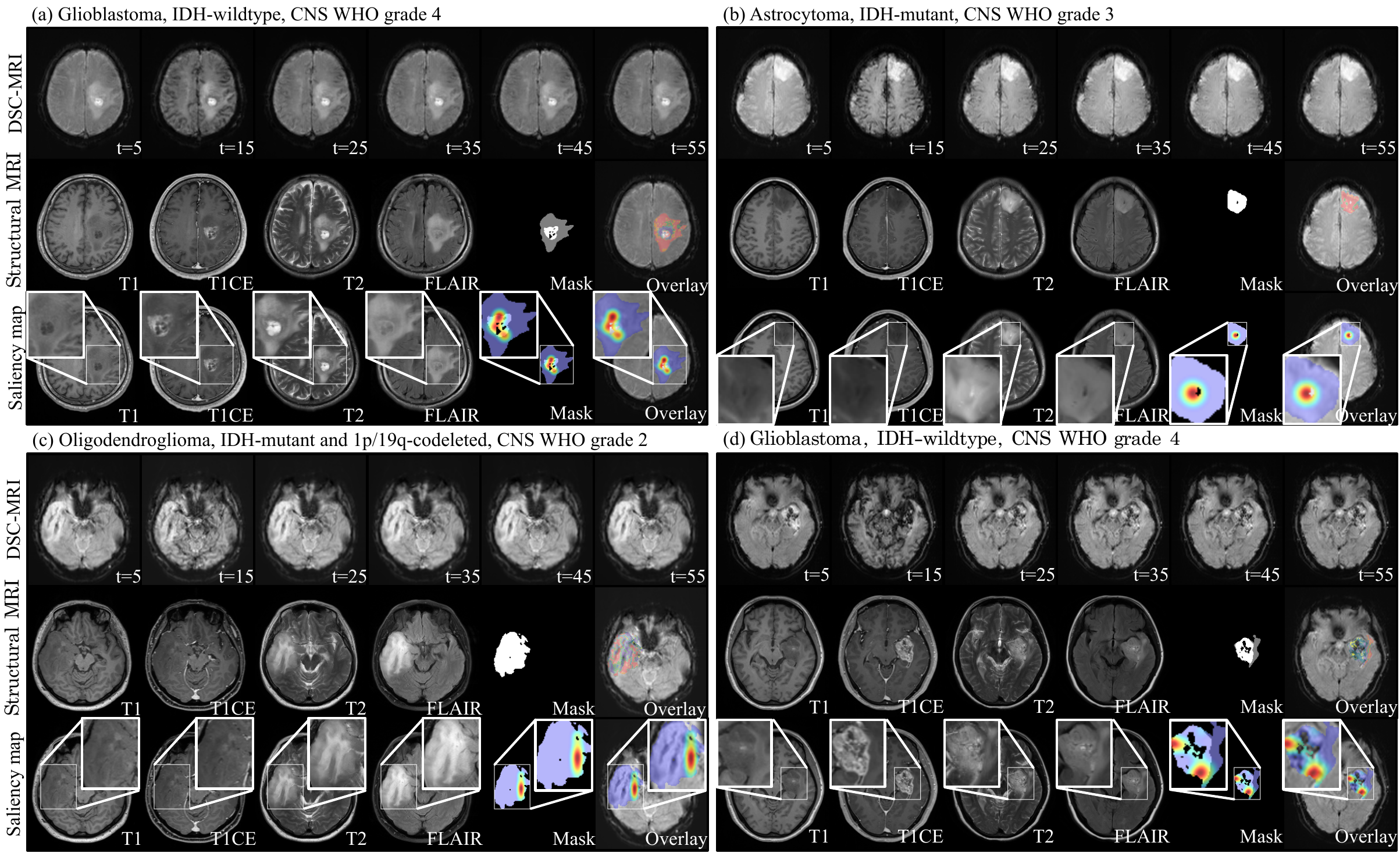}
\caption{
\textbf{Saliency-guided interpretability across molecular subgroups.}
Rows: DSC perfusion frames, structural MRI with perfusion code overlay, saliency maps.
(a)~GBM, IDH-wildtype, Grade~4 (66M).
(b)~Astrocytoma, IDH-mutant, Grade~3 (41F).
(c)~Oligodendroglioma, IDH-mutant, 1p/19q-codeleted, Grade~2 (49F).
(d)~GBM, IDH-wildtype, Grade~4 (53F).
}
\label{fig:saliency}
\end{figure}

\subsection{Model Interpretability}

Figure~\ref{fig:saliency} presents gradient-based saliency maps.
In Case~(a), an IDH-wildtype glioblastoma, high attribution concentrates on the heterogeneously enhancing rim and infiltrative margins surrounding central necrosis, consistent with the hypervascular phenotype of IDH-wildtype tumors.
Conversely, Case~(b), an IDH-mutant astrocytoma, shows distributed importance across non-enhancing infiltrative components on T2/FLAIR rather than focal enhancement, reflecting less angiogenic biology.
Case~(c), an oligodendroglioma with 1p/19q codeletion, highlights clustered perfusion codes along the lesion and perilesional tissue, capturing the heterogeneous texture typical of this subtype.
Case~(d), another IDH-wildtype glioblastoma, emphasizes both the necrotic rim and patchy perfusion hotspots driving classification.

Across cases, the model consistently assigns high attribution to regions reflecting known genotype-specific vascular biology, with IDH-wildtype tumors activating enhancing and hypervascular zones while IDH-mutant tumors shift attention toward infiltrative, non-enhancing components.
Quantitatively, the model assigned over 12-fold higher saliency to the enhancing tumor region in IDH-wildtype cases than in IDH-mutant cases~($p{<}0.001$), confirming that predictions are driven by angiogenic signatures aligned with known pathophysiology.

\section{Conclusion}

We presented HiPerfGNN, a radiogenomic framework that integrates unsupervised perfusion quantization with structure-guided hierarchical graph construction for glioma molecular subtyping.
By encoding raw DSC-MRI time-intensity curves into discrete representations of hemodynamic patterns via VQ-VAE and organizing them into tumor-specific multi-scale graphs, the model captures complementary functional and anatomical information.
Our approach achieved strong performance for IDH mutation, 1p/19q codeletion, and WHO grade prediction, and demonstrated robust cross-site generalization on an independent external cohort without recalibration.
Gradient-based saliency analysis confirmed that the model attends to biologically meaningful tumor subregions.
This work demonstrates a scalable pathway for integrating dynamic perfusion imaging into precision oncology.
Future work will extend validation to multi-center prospective cohorts, explore joint multi-task learning across molecular targets, and investigate contrast-free alternatives such as arterial spin labeling~(ASL) MRI.

\begin{credits}
\subsubsection{\ackname}
This work was supported by the National Research Foundation of Korea (NRF) grant funded by the Korea government (MSIT) (No. RS-2026-25479661) (K.S.C.); the SNUH Research Fund (No. 04-2024-0600; No. 04-2025-2060) (K.S.C.); and the Korea Health Technology R\&D Project through the Korea Health Industry Development Institute (KHIDI) grant funded by the Ministry of Health\&Welfare (No. RS-2024-00439549) (K.S.C.).

\subsubsection{\discintname}
The authors have no competing interests to declare that are relevant 
to the content of this article.
\end{credits}

\bibliographystyle{splncs04}
\bibliography{ref}

\end{document}